\documentclass[lettersize,journal]{IEEEtran}
\usepackage{amsmath,amsfonts}
\usepackage{algorithmic}
\usepackage{algorithm}
\usepackage {ulem}

\usepackage{array}
\usepackage[caption=false,font=normalsize,labelfont=sf,textfont=sf]{subfig}
\usepackage{textcomp}
\usepackage{stfloats}
\usepackage{url}
\usepackage{verbatim}
\usepackage{svg}
 
\usepackage{graphicx}
\usepackage{booktabs}
\usepackage{color,balance, cite, bm, xcolor} 
\usepackage{amssymb}
\usepackage{hyperref}
\hypersetup{hidelinks}
\usepackage{threeparttable}
\usepackage{multirow}
\usepackage{multicol}

\def \eg {\emph{e.g.}}
\def \ie {\emph{i.e.}}
\def \etal {\emph{et al. }}

\begin{document}


\title{LP\textsuperscript{2}DH: A Locality-Preserving Pixel-Difference Hashing Framework for Dynamic Texture Recognition}

\author{Ruxin Ding,~\IEEEmembership{Student Member,~IEEE,}
Jianfeng Ren,~\IEEEmembership{Senior Member,~IEEE,}
Heng Yu,~\IEEEmembership{Member,~IEEE,}
Jiawei Li,~\IEEEmembership{Member,~IEEE,}
Xudong Jiang,~\IEEEmembership{Fellow,~IEEE}
\thanks{R. Ding, J. Ren, H. Yu and J. Li are with the School of Computer Science, University of Nottingham Ningbo China, Ningbo 315100, China. 
	
X. Jiang is with School of Electrical \& Electronic Engineering, Nanyang Technological University, Nanyang Link, 639798 Singapore. 

\textit{Corresponding author: E-mail: 
	\href{mailto:jianfeng.ren@nottingham.edu.cn}{jianfeng.ren@nottingham.edu.cn}}.
    
This work was supported by Ningbo Municipal Bureau of Science and Technology under Grant 2023Z138, 2023Z237,  2024Z110 and 2024Z124. 
}}

\markboth{IEEE TRANSACTIONS ON IMAGE PROCESSING}%
{Shell \MakeLowercase{\textit{et al.}}: A Sample Article Using IEEEtran.cls for IEEE Journals}


\maketitle

\begin{abstract}
Spatiotemporal Local Binary Pattern (STLBP) is a widely used dynamic texture descriptor, but it suffers from extremely high dimensionality. To tackle this, STLBP features are often extracted on three orthogonal planes, which sacrifice inter-plane correlation. In this work, we propose a Locality-Preserving Pixel-Difference Hashing (LP\textsuperscript{2}DH) framework that jointly encodes pixel differences in the full spatiotemporal neighbourhood. LP\textsuperscript{2}DH transforms Pixel-Difference Vectors (PDVs) into compact binary codes with maximal discriminative power. Furthermore, we incorporate a locality-preserving embedding to maintain the PDVs’ local structure before and after hashing. Then, a curvilinear search strategy is utilized to jointly optimize the hashing matrix and binary codes via gradient descent on the Stiefel manifold. After hashing, dictionary learning is applied to encode the binary vectors into codewords, and the resulting histogram is utilized as the final feature representation. The proposed LP\textsuperscript{2}DH achieves state-of-the-art performance on three major dynamic texture recognition benchmarks: 99.80\% against DT-GoogleNet's 98.93\% on UCLA, 98.52\% against HoGF\textsuperscript{3D}’s 97.63\% on DynTex++, and 96.19\% compared to STS's 95.00\% on YUPENN. The source code is available at: \url{https://github.com/drx770/LP2DH}. 
\end{abstract}


\section{Introduction}

\IEEEPARstart{D}{ynamic} texture (DT) refers to sequences of images that consist of repeated patterns related to time and space~\cite{nguyen2021prominent}. 
DT recognition has been widely used in various applications, \eg, video anomaly detection~\cite{hu2019squirrelcage}, fire detection~\cite{ding2024discriminant}, 
facial expression recognition~\cite{poux2022dynamic},  face spoofing detection~\cite{zhao2018dynamic}, and dynamic scene recognition~\cite{derpanis2012dynamic}. 
Compared to static textures~\cite{pan2024l0,upadhyay2025sarlbp}, recognizing DTs is more challenging because their appearance changes over time~\cite{zhao2019dynamic}. Therefore, a descriptive feature representation is the key to effective DT recognition.

A wide range of methods have been developed for dynamic texture recognition~\cite{nguyen2021comprehensive}. These include: 1)~optical flow-based approaches that capture temporal variations~\cite{chen2023optical,wan2024event,poux2022dynamic,nguyen2018directional}; 
2)~methods modelling dynamics as linear dynamical systems~\cite{wang2016dynamic,wei2017dynamical,mumtaz2012clustering,xing2023dynamic}; 
3)~fractal geometry-based models~\cite{xu2015classifying, quan2017spatiotemporal}; 
4)~spatiotemporal filtering techniques to tackle noise and illumination variations~\cite{zhao2019dynamic, arashloo2014dynamic, nguyen2021prominent}; 
5)~local spatiotemporal feature encoding methods such as STLBP~\cite{zhao2018dynamic,hu2019squirrelcage} and other local descriptors~\cite{nguyen2020rubik, nguyen2020momental, zhao2023dynamic}; 
and 6)~deep learning~\cite{andrearczyk2018convolutional, gan2018geometry, hadji2017spatiotemporal, wang2022self} and representation learning approaches~\cite{quan2015dynamic, quan2016equiangular, junior2019randomized, ding2022dynamic}. Among these, STLBP descriptors~\cite{zhao2018dynamic, hu2019squirrelcage} are particularly popular for their computational simplicity, robustness to illumination, and effective spatiotemporal encoding.

Despite progress, STLBP descriptors still face three key challenges. 1)~High dimensionality: Descriptors like VLBP~\cite{zhao2007dynamic} grow exponentially with neighbourhood size $P$ (\eg, a $P\times P\times P$ neighbourhood yields $2^{P^3-1}$ possible codes), making them impractically high-dimensional. 
2)~Suboptimal encoding: Conventional LBP uses simple thresholding, which fails to capture rich context from pixel differences~\cite{ren2013noise, lu2015learning, ren2015lbp, duan2018contextaware}. 
3)~Lost structural information: Mapping continuous differences to binary codes can break important relational structure, reducing discriminative power. These limitations collectively highlight the need for a more advanced encoding approach.

To address the high-dimensionality problem, we propose a Locality-Preserving Pixel-Difference Hashing (LP\textsuperscript{2}DH) framework that produces compact binary descriptors. Previous methods have specific drawbacks: LBP-TOP reduces dimensionality but sacrifices inter-plane correlations~\cite{zhao2007dynamic}; structural optimization of LBP loses texture details~\cite{ren2015learning}; and straightforward dictionary learning struggles with very high-dimensional features~\cite{ding2022dynamic}. Our LP\textsuperscript{2}DH overcomes these issues with a two-stage design. First, we perform Pixel-Difference Vector (PDV) hashing to map a high-dimensional PDV (\eg, a $5\times 5\times 5$ neighbourhood, 124-dimensional) into a much lower-dimensional binary code (\eg, $M=40$) while preserving important micro-structural information. Second, we apply dictionary learning to cluster the resulting binary vectors, reducing the feature dimension from an initial $2^M$ possible patterns down to $C$ representative codewords. This two-stage strategy greatly reduces dimensionality while maintaining discriminative power, yielding a compact yet representative DT descriptor.

To tackle the challenge of optimal pixel-difference encoding, we propose a multi-objective PDV hashing scheme that systematically optimizes the encoding process. Existing approaches improve certain aspects of LBP encoding, but they do not provide a unified objective. For example, noise-resistant variants such as uniform LBP enhance robustness by imposing quantization constraints~\cite{ren2013noise}, and SARLBP incorporates multi-scale texture information~\cite{upadhyay2025sarlbp}. More comprehensive descriptors such as completed VLBP~\cite{tiwari2016dynamic} and multi-channel LBP~\cite{ren2017sound} further exploit both sign and magnitude components. However, these techniques typically optimize only part of the encoding pipeline and lack a formulation that jointly accounts for information preservation, discriminability, and mapping accuracy. In contrast, our framework simultaneously 1) minimizes quantization loss to improve mapping accuracy, 2) maximizes binary-code entropy to preserve information content, and 3) maximizes code variance to enhance discriminative power. This integrated formulation yields a better-balanced and more discriminative binary representation of pixel differences.

Lastly, to preserve local structural relationships during binary mapping, we incorporate a locality-preserving criterion directly into the PDV hashing formulation. Although many manifold learning methods aim to preserve locality, they are rarely integrated into hashing-based feature extraction pipelines for dynamic texture recognition. Inspired by Locally Linear Embedding~\cite{ji2017toward}, we enforce that PDVs that are close in the original space should remain close after hashing, \ie, similar PDVs should yield proximal binary codes. We integrate this as one of the four optimization objectives in our hashing scheme. This ensures that neighbourhood relationships are maintained throughout the transformation. Hence, the intrinsic local structure of dynamic texture patterns is better preserved.

Our main contributions are summarized as follows. 
1)~We propose LP\textsuperscript{2}DH, which jointly optimizes hashing functions and binary codes via gradient descent on the Stiefel manifold, providing an effective solution to the high-dimensionality issue of STLBPs.  
2)~We develop a PDV hashing scheme that integrates a locality-preserving criterion with objectives of minimizing quantization loss and maximizing code entropy and variance, enabling more informative and discriminative encoding of PDVs. 
3)~We demonstrate that LP\textsuperscript{2}DH achieves state-of-the-art performance on three benchmark datasets for dynamic texture recognition.

\section{Related work}

\subsection{Dynamic Texture Recognition}
Dynamic texture recognition methods are broadly categorized as follows~\cite{nguyen2021comprehensive}. 

\noindent\textbf{1)~Optical-flow-based methods} emphasize the motion of objects across adjacent frames~\cite{nguyen2018directional}. For instance, Feature of Directional Trajectory (FDT)~\cite{nguyen2018directional} captures dominant motion orientations by computing histograms of optical-flow vectors along dense trajectories. Its extension, FD-MAP~\cite{nguyen2018directional}, further encodes motion curvature by quantizing turning angles between successive segments. While these methods effectively capture distinctive motion patterns, they are computationally expensive, sensitive to estimation errors in complex scenes, and often ignore static visual information. 

\noindent\textbf{2)~Model-based methods} characterize DTs by employing models like Linear Dynamical Systems (LDS) to represent temporal dynamics. Wei~\etal developed a Joint Video Dictionary Learning (JVDL) framework, which learns a linear transition matrix between sparse coefficients of consecutive frames for synthesis and recognition~\cite{wei2017dynamical}. Mumtaz~\etal developed a hierarchical EM-based clustering algorithm that learns cluster centers consistent with the generative model, supported by recursive Kalman smoothing~\cite{mumtaz2012clustering}. Moreover, Xing~\etal introduced a bag-of-models approach using a mixture of Switching Hidden Markov Models (SHMMs), where EM-derived components form a codebook for classification~\cite{xing2023dynamic}. These approaches offer principled temporal modelling but are limited by assumptions of linearity and Gaussian noise, which may not generalize well to complex real-world dynamics. 

\noindent\textbf{3)~Geometry-based methods} characterize DTs using fractal geometry principles. Quan~\etal developed a lacunarity-based descriptor to encode spatio-temporal irregularities in LBP features, providing robustness to illumination and viewpoint changes~\cite{quan2017spatiotemporal}. Xu~\etal developed a dynamic fractal analysis framework that characterizes stochastic self-similarity using a volumetric dynamic fractal spectrum~\cite{xu2015classifying}. While offering illumination and viewpoint invariance through scale-invariant analysis, these methods are sensitive to noise and can degrade on videos with weak fractal properties or strong stochasticity. 

\noindent\textbf{4)~Filter-based methods} leverage spatiotemporal filtering to extract robust features. Among them, B3DF\_SMC~\cite{zhao2019dynamic} is an unsupervised 3D filter learning method, which generates spatiotemporal binary features by jointly histogramming the sign and magnitude components of filter responses~\cite{zhao2019dynamic}. Nguyen~\etal employed multi-scale, high-order Gaussian-gradient filtering followed by shallow encoding to construct discriminative HOGF\textsuperscript{2D/3D} descriptors~\cite{nguyen2021prominent}. Rivera and Chae introduced the Dynamic Number transitional Graph (DNG) descriptor, which computes directional transitions between frames to form graph signatures, then aggregates them into a sequence-level representation~\cite{rivera2015spatiotemporal}. MBSIF-TOP employs ICA-learned filters on three orthogonal planes for binary code generation and uses a multi-resolution scheme for multi-scale representation~\cite{arashloo2014dynamic}. These methods encode robust local statistics through filter banks but often struggle to capture complex motions and long-term dependencies due to inherent filter constraints. 

\noindent\textbf{5)~Local-feature-based methods} characterize DTs by computing statistics from local spatiotemporal neighbourhoods. Early methods like VLBP often suffer from high dimensionality due to exponential feature growth~\cite{zhao2007dynamic}. LBP-TOP alleviates this by extracting features from three orthogonal planes~\cite{zhao2007dynamic}. VLBC/CVLBC replaces binary encoding with pixel counting to avoid dimensionality explosion while integrating motion and appearance cues~\cite{zhao2018dynamic}. Ren~\etal optimized LBP structure first via an incremental feature selection method before using it for histogram feature generation~\cite{ren2015learning}. Nguyen~\etal developed Local Rubik-based Pattern to jointly capture shape and motion~\cite{nguyen2020rubik}, and Momental Directional Patterns to hybridize filtering and local-feature advantages~\cite{nguyen2020momental}. Directional Binarized Random Features (DBRFs) further extract Gaussian-gradient-based directional information from DTs~\cite{zhao2023dynamic}. Despite computational efficiency and illumination robustness, these methods face a persistent trade-off between discriminative power and feature compactness in their spatiotemporal extensions. 

\noindent\textbf{6)~Learning-based methods} are divided into those using handcrafted features and deep learning approaches. Handcrafted-feature methods include: Orthogonal Tensor Dictionary Learning (OTDL)~\cite{quan2015dynamic} and Equiangular Kernel Dictionary Learning (EKDL)~\cite{quan2016equiangular}, which extract features directly from video tensors; dictionary learning applied to motion features from Slow Feature Analysis (SFA)~\cite{theriault2013dynamic}; and PI-LBP~\cite{ren2013dynamic}, which leverages Principal Histogram Analysis and super histograms for robust LBP representation learning. 
Initial deep learning approaches for DT recognition, including AlexNet and DT-GoogleNet~\cite{andrearczyk2018convolutional}, were primarily supervised. Subsequently, research shifted towards self-supervised learning to leverage unlabeled data. In this paradigm, methods like Unsupervised-CNN~\cite{wang2015unsupervised} and ClipOrder~\cite{misra2016shuffle} both employ a Siamese-triplet network architecture built upon AlexNet. They differ, however, in their designed auxiliary tasks: the former uses a ranking loss for feature embedding, while the latter learns by verifying the temporal order of video clips. 
Alternatively, Gan~\etal utilized a FlowNet Simple backbone with geometric consistency as the self-supervisory signal~\cite{gan2018geometry}. 
Recently, Wang~\etal developed Spatiotemporal Statistics (STS)~\cite{wang2022self}, which extracts motion and appearance statistics as auxiliary signals to train 3D networks in a self-supervised manner. Learning-based methods, especially deep networks, automatically learn complex spatiotemporal features and often outperform handcrafted methods, but at the cost of high computational demand, potential overfitting on small datasets, and lower interpretability.

\subsection{Spatiotemporal Local Binary Patterns}
STLBPs are widely used for dynamic texture (DT) recognition due to their simplicity and robustness to illumination changes~\cite{zhao2007dynamic,hu2019squirrelcage,huang2016spontaneous,ren2017sound,zhao2018dynamic}. The paradigm originated with VLBP~\cite{zhao2007dynamic}, which encodes pixel differences in a cubic neighbourhood but suffers from a feature dimension that grows exponentially as \(2^{P^3-1}\), severely limiting its practical use.

Subsequent research has thus developed several categories of methods to mitigate this issue. 
1)~\textbf{LBP-TOP and variants}~\cite{zhao2007dynamic,arashloo2014dynamic,chai2024blind} reduce dimensionality and computational cost by extracting features from three orthogonal planes. However, this simplification discards the potentially useful spatiotemporal correlations between planes. 
2)~\textbf{Uniform LBPs}~\cite{singh2018color,zeebaree2021multi,jia2017local} achieve feature compactness by treating infrequent non-uniform patterns as noise and consolidating them into a single histogram bin. While this enhances robustness, it risks discarding rare but discriminative patterns, potentially reducing recognition accuracy. 
3)~\textbf{LBP structure optimization methods}~\cite{ren2017lbp,ren2014optimizing,ren2015learning} reduce dimensionality by optimizing the LBP operator's neighbour configuration. For example, Ren~\etal solved the neighbour selection problem using an incremental scheme by maximizing conditional mutual information~\cite{ren2015learning} and via binary quadratic programming~\cite{ren2014optimizing}. These methods yield compact, information-preserving codes but are computationally intensive and may overfit to the training data. 
4)~\textbf{Dimensionality reduction methods}, such as Principal Component Analysis~\cite{ding2024discriminant, ren2013dynamic,ren2015chi}, Locality Preserving Projections~\cite{chao2015facial}, and Laplacian Eigenmaps~\cite{guerrero2017group}, project high-dimensional LBP features into a compact, lower-dimensional subspace. While effective for storage and generalization, these linear or manifold-based projections can lose discriminative details and reduce feature interpretability.
5)~\textbf{Dictionary-learning methods} learn a compact dictionary of representative codewords from the high-dimensional feature space~\cite{lu2015joint, huang2016spontaneous, duan2018contextaware}. For instance, Huang~\etal learned discriminative codebooks to encode and fuse sign, magnitude, and orientation components of local patterns~\cite{huang2016spontaneous}. Duan~\etal constructed the codebook by clustering binary codes derived from PDVs~\cite{duan2018contextaware}. Although they achieve compact and invariant representations, their performance hinges on clustering quality and incurs significant optimization overhead. 

Another major challenge is that traditional LBPs typically encode only the sign of pixel differences, ignoring the potentially informative magnitude~\cite{zhao2007dynamic,ren2017sound}. To address this, several strategies have been developed:
1)~\textbf{Ternary thresholding methods} such as Squirrel-Cage LBP (SCLBP)~\cite{hu2019squirrelcage} encode magnitude information using a third state, creating ternary codes. This improves robustness to noise and illumination but introduces the challenge of optimal threshold selection and increased feature complexity.
2)~\textbf{Multi-component encoding methods}~\cite{huang2016spontaneous} utilize multiple components of PDVs (\eg, sign, magnitude, orientation) separately to build a richer representation. While more descriptive, they require careful management to avoid a combinatorial explosion in feature dimensionality. 
3)~\textbf{Joint encoding methods}~\cite{ren2013noise,lu2015learning,duan2018contextaware} jointly encode both sign and magnitude into a unified, compact binary descriptor. For instance, Ren~\etal developed a noise-resistant LBP that forms uniform codes by jointly considering the central bit and its neighbours~\cite{ren2013noise}. Compact Binary Face Descriptor~\cite{lu2015learning} and CA-LBMFL~\cite{duan2018contextaware} utilize learning frameworks to achieve this integration. These methods yield descriptors that are robust to noise and illumination, albeit with increased design complexity over sign-only approaches.

\section{Proposed LP\textsuperscript{2}DH}
\subsection{Preliminaries of Pixel Difference Hashing}
To overcome the limitations of high dimensionality, fragile encoding, and structural information loss, we introduce a pixel-difference vector (PDV) hashing framework.
For each video, a local spatiotemporal neighbourhood of size $P\times P\times P$ is extracted. Let $I_c$ denote the intensity of the center pixel and $I_p$ the intensity of the $p$-th neighbouring pixel, where $p=1,2,\ldots, P^3-1$. The PDV is defined as $\bm{x}=[I_1-I_c,I_2-I_c,\ldots,I_D-I_c]^{\top}\in \mathbb{R}^{D}$, where $D = P^{3}-1$. Collecting all $N$ PDVs extracted from the video yields the PDV matrix $\bm{X} = [\bm{x}_1, \ldots, \bm{x}_N] \in \mathbb{R}^{D \times N}$.

The objective is to learn a hashing projection matrix $\bm{W} = [\bm{w}_1, \bm{w}_2, \ldots, \bm{w}_M]\in \mathbb{R}^{D \times M}$ that maps each PDV $\bm{x}_n$ to an $M$-bit binary code $\bm{b}_n \in \{0,1\}^{M}$. Each bit is generated as, 
\begin{equation}
    b_{mn} = f_s(\bm{w}_m^{\top}\bm{x}_n),
    \label{eq:bmn}
\end{equation}
where $\bm{w}_m \in \mathbb{R}^{D}$ is the projection vector for the $m$-th hashing function. The thresholding function $f_s(v)$ equals $1$ if $v \ge 0$ and $0$ otherwise. The complete output is the binary code matrix $\bm{B} = [\bm{b}_1, \bm{b}_2, \ldots, \bm{b}_N]\in \{0,1\}^{M\times N}$, which serves as the basis for constructing histogram-based features.

\begin{figure*}[htbp] 
	\centering
	\includegraphics[width=0.92\textwidth]{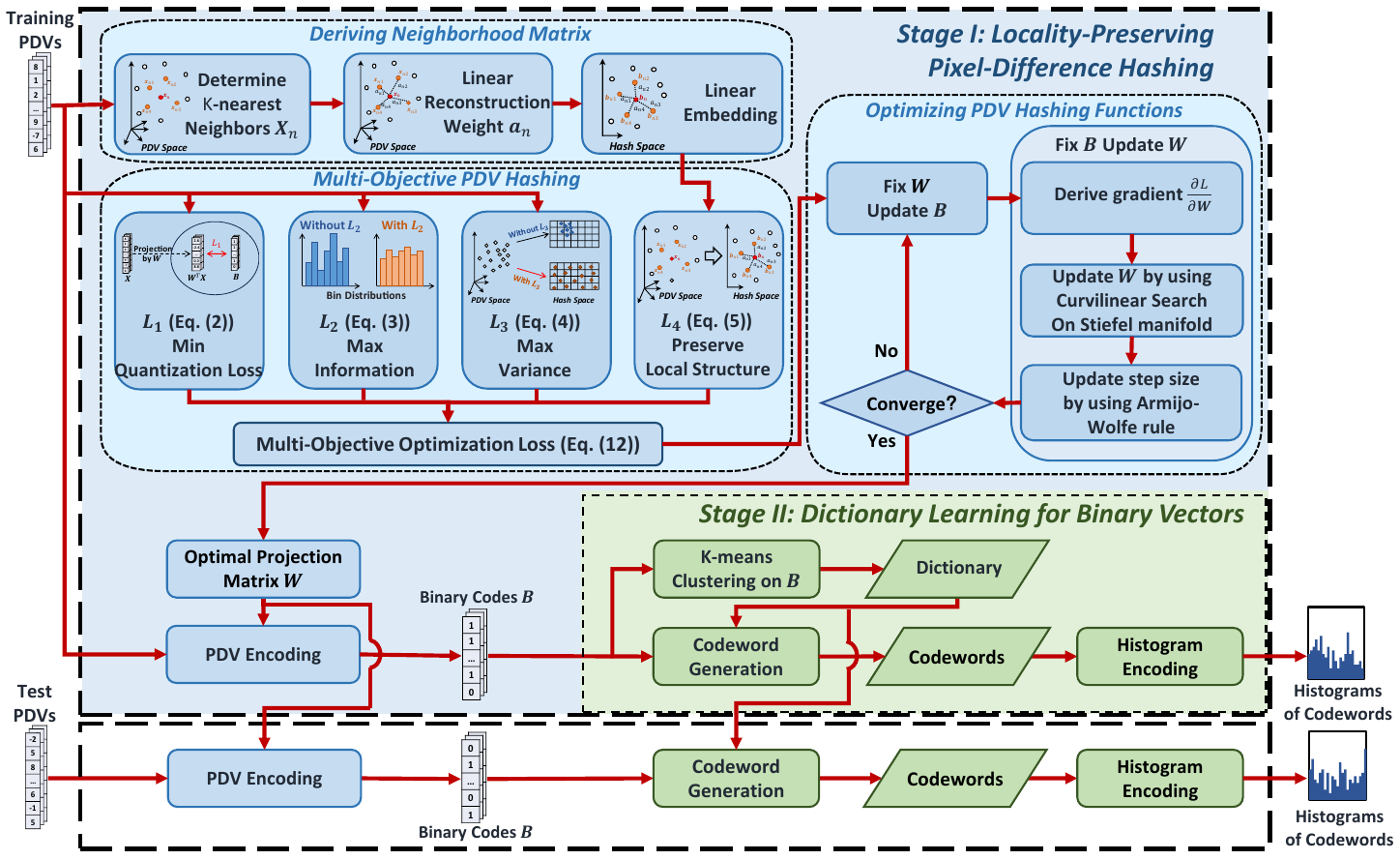}
	\caption{Overview of proposed LP\textsuperscript{2}DH. It derives compact and discriminative features in two stages: 1)~Locality-Preserving Pixel Difference Hashing, which maps PDVs to binary codes using an optimized projection matrix $\bm{W}$. The matrix is learned through a multi-objective formulation that minimizes quantization loss ($L_1$), maximizes information ($L_2$) and variance ($L_3$) and preserves local topology ($L_4$), solved via Stiefel manifold optimization. 2)~Dictionary Learning for Binary Vectors, which clusters binary codes into visual words and encodes them into a histogram-of-codewords representation for classification.} 
\label{fig:block_diagram}
\end{figure*}

\subsection{Overview of Proposed LP\textsuperscript{2}DH}
As illustrated in Fig.~\ref{fig:block_diagram}, we propose the Locality-Preserving Pixel-Difference Hashing (LP\textsuperscript{2}DH) framework for PDV feature encoding, which consists of two main stages. 

\noindent\textbf{PDV hashing Module:} \quad
This module maps each PDV of size $P^3-1$ into a compact $M$-bit binary code via learned hash functions, by jointly encoding both sign and magnitude components. The hash functions are optimized to satisfy four key criteria: minimizing quantization loss, maximizing information retention, maximizing code variance for compactness, and preserving local data structure. The first two are standard objectives in feature hashing to ensure fidelity~\cite{wang2022learning}, while the latter two specifically enhance the representation: variance maximization promotes separable codes, and locality preservation ensures similar PDVs yield similar binary codes. A gradient-descent method on the Stiefel manifold is then applied to alternately optimize the hash functions and the binary codes with respect to this combined objective.

\noindent\textbf{Dictionary-learning Module:} \quad 
The $M$-bit binary codes are clustered to form a dictionary of $C$ codewords, which is then used to encode the data into a compact histogram representation. This dramatically reduces the dimensionality from $2^M$ to $C$. Finally, multi-scale features are obtained by extracting LP\textsuperscript{2}DH descriptors from neighbourhoods of different sizes.

\subsection{PDV Hashing Module}

We formulate PDV hashing as a multi-objective optimization problem with four loss terms: a quantization loss to minimize binarization error, an information loss to balance bit usage, a variance loss to enhance discriminability, and a locality-preserving loss to maintain neighbourhood relationships. The details of these objectives are presented below.

\noindent \textbf{Minimize Quantization Loss:} \quad
Quantization of PDVs into binary codes will lead to a quantization loss. We hence aim to minimize the following quantization loss, 
\begin{align}
\label{eqn:quantiztion_loss}
L_1 = \sum_{n=1}^{N} \sum_{m=1}^{M} {( b_{mn}-\bm{w}_{m}^{\top}\bm{x}_n )^2}.
\end{align}

\noindent \textbf{Maximize Information of Hash Codes:} \quad
This objective maximizes the entropy of the hash codes by promoting a uniform distribution across all possible codewords: 
\begin{align}
\label{eqn:max_info}
L_2 &= \sum_{m=1}^{M} {( \sum_{n=1}^{N} (b_{mn} - 0.5) )^2}. 
\end{align}
Minimizing $L_2$ ensures that each bit approaches equal probability of being 0 or 1, where the constant 0.5 represents the midpoint of the binary domain and promotes maximum code entropy. Any deviation from this value would bias the distribution and reduce information capacity \cite{zhang2022probabilistic,wang2017survey}.  

\noindent \textbf{Maximize Variance of Hash Codes:} \quad
This objective enhances code compactness by promoting balanced bit usage and reducing representational redundancy, \ie, 
\begin{align}
\label{eqn:L4}
L_3 &= -\sum_{n=1}^{N} \sum_{m=1}^{M} {( b_{mn} - \mu_m )^2},   \end{align}
where $\mu_m$ is the mean of the $m$-th bit of all $N$ PDVs. Maximizing $L_3$ increases bit-wise variance, helping eliminate redundancy and yielding a more compact representation~\cite{wang2022learning}. 

\noindent \textbf{Preserve Local Data Structure:} \quad
To maintain the intrinsic local geometry of the data during hashing, we enforce that neighbouring relations among PDVs in the original high-dimensional space are preserved in the resulting binary code space. Given PDV matrix $\bm{X}$ in the original space, let \( \bm{A} \in \mathbb{R}^{N \times N} \) be a neighbourhood affinity matrix, where each element \( a_{nk} \) encodes the neighbouring relationship between \( \bm{x}_n \) and \( \bm{x}_k \). This matrix is constructed using Locally Linear Embedding (LLE)~\cite{ji2017toward,he2023learning} to capture the local linear structure of the data; details are provided in Section~\ref{part:LLE}.

The same neighbourhood structure is then imposed in the binary code space by minimizing the following loss:
\begin{align}
\label{eqn:J1}
L_4 &= \sum_{n=1}^N \Big\Vert \bm{b}_n - \sum_{k=1}^N a_{nk} \bm{b}_k \Big\Vert^2.
\end{align}
Intuitively, minimizing \( L_4 \) encourages each binary code \( \bm{b}_n \) to be reconstructible as a linear combination of its neighbours, with weights given by \( \bm{A} \). This ensures that local proximity in the original PDV space is reflected in the binary representation.

\noindent \textbf{Final Loss Function:} \quad
By integrating the four loss terms, the final objective function for PDV hashing is formulated as, 
\begin{equation}
\begin{aligned}
\label{eqn:objective}
    L= \lambda_1L_1 + \lambda_2L_2 + \lambda_3L_3 + L_4, 
\end{aligned}
\end{equation}
where $\lambda_1$, $\lambda_2$ and $\lambda_3$ control the trade-off among different objectives. 
The goal is to learn the projection matrix $\bm{W}$ that minimizes the loss $L$, thereby yielding compact, balanced, and discriminative binary codes while preserving local structure.

\subsection{Deriving the Neighbourhood Matrix} 
\label{part:LLE}
To preserve local structure during hashing, we resort to Locally Linear Embedding~\cite{he2023learning}. LLE models each PDV as a linear combination of its nearest neighbours in the original space. By enforcing the same reconstruction relationships in the binary code space, the local topology is preserved after mapping. The detailed procedure is as follows:

\noindent\textbf{$K$-Nearest Neighbour Identification:} \quad
For each PDV $\bm{x}_n$, we first identify its $K$ nearest neighbours. Let $\bm{x}_{nk}$ denote the $k$-th neighbour and $\bm{X}_n = [\bm{x}_{n1},\bm{x}_{n2},\dots,\bm{x}_{nK}] \in \mathbb{R}^{D \times K}$ be the matrix of all $K$ neighbours of $\bm{x}_n$.

\noindent\textbf{Linear Reconstruction}: \quad  
We then derive the optimal weights $\Tilde{\bm{a}}_{n}$ for a linear reconstruction of $\bm{x}_n$ as,
\begin{equation}
\begin{aligned}
    \label{eqn:LLE}    
    \min_{\{\Tilde{\bm{a}}_{n}\}_{n=1}^N}
    & \sum_{n=1}^{N} \Vert{ {\bm{x}_n}-\sum_{k=1}^{K} \Tilde{a}_{nk}\bm{x}_{nk} }\Vert^2\\ 
    &s.t. \sum_{k=1}^K \Tilde{a}_{nk}=1, \forall{n} \in \{1,\dots,N\}, 
\end{aligned}
\end{equation}
where $\Tilde{\bm{a}}_n =[\Tilde{a}_{n1},\Tilde{a}_{n2},\dots,\Tilde{a}_{nK}]^{\top}\in\mathbb{R}^{K}$ denotes the reconstruction weights for $\bm{x}_n$ using its $K$ neighbours $\{\bm{x}_{nk}\}_{k=1}^K$.
Eq.~\eqref{eqn:LLE} can be rewritten as,
\begin{equation}
\label{eqn:maGa}
\begin{aligned}
\min_{\{\Tilde{\bm{a}}_{n}\}_{n=1}^N} &\sum_{n=1}^{N}\Tilde{\bm{a}}_{n}^{\top} \bm{G}_n \Tilde{\bm{a}}_{n}, \\
&s.t. \quad  \bm{1}_K^{\top}\Tilde{\bm{a}}_{n}=1, \forall{n} \in \{1,2,\dots, N\}, 
\end{aligned}
\end{equation}
where $\bm{1}_K = [1,1, \dots,1]^{\top} \in \mathbb{R}^{K}$, $\bm{G}_n = (\bm{x}_n\bm{1}_K^{\top}-\bm{X}_n)^{\top}(\bm{x}_n\bm{1}_K^{\top}-\bm{X}_n) \in \mathbb{R}^{K\times K}$. 
The optimization problem in Eq.~\eqref{eqn:maGa} can be solved using the Lagrangian multiplier method~\cite{li2023self}:  
\begin{equation}
\begin{aligned}
    \label{eqn:LLE_Lagrangian}
\mathcal{L} = \sum_{n=1}^{N}\Tilde{\bm{a}}_{n}^{\top} \bm{G}_i \Tilde{\bm{a}}_{n} - \sum_{n=1}^{N} \lambda_n (\bm{1}_K^{\top}\Tilde{\bm{a}}_{n}-1), 
\end{aligned}
\end{equation} 
where $\lambda_n$ is the Lagrange multiplier. 
Optimal $\Tilde{\bm{a}}_{n}$ is derived as,
\begin{equation}
\begin{aligned}
    \label{eqn:LLE_Lagrangian5}
\Tilde{\bm{a}}_{n} = \frac{\bm{G}_n^{-1}\bm{1}_K}{\bm{1}_K^{\top}\bm{G}_n^{-1}\bm{1}_K}.
\end{aligned}
\end{equation}

\noindent\textbf{Linear Embedding}: \quad
After deriving the weights $\{\Tilde{\bm{a}}_{n}\}_{n=1}^N$, the same local structure is enforced in the low-dimensional space of binary vectors by minimizing the following loss,
\begin{align}
    \label{eqn:LLE_reconstruct}
   L_{E}=\sum_{n=1}^N \Vert{\bm{b}_n-\sum_{k=1}^N a_{nk} \bm{b}_k} \Vert^2,
\end{align}
where $a_{nk} = \Tilde{a}_{n\Tilde{k}}$ if $\bm{x}_k$ is
the $\Tilde{k}$-th neighbour of $\bm{x}_n$, and zero otherwise. 
$L_{E}$ contributes to the fourth loss term in Eq.~\eqref{eqn:objective}.

\subsection{Optimizing PDV Hashing Functions}
\label{sec:solve}

To jointly optimize the projection vectors $\bm{w}_m$, $\forall m \in\{1,2,\dots,M\}$, we rewrite the loss function in Eq.~\eqref{eqn:objective} as,
\begin{equation}
\begin{aligned}
\label{eqn:objective2}
    L &= \lambda_1L_1 + \lambda_2L_2 + \lambda_3L_3 + L_4 \\  
    &= \lambda_1 {\Vert \bm{B}  - \bm{W}^{\top}\bm{X} \Vert^2} + \lambda_2 {\Vert (\bm{B} - 0.5)\times\bm{1}_N \Vert^2} \\
    &- \lambda_3\Vert\bm{B}-\bm{U}\Vert^2 + \Vert{\bm{B}-\bm{B}\bm{A}}\Vert^2,
\end{aligned}
\end{equation}
where $\bm{1}_N = [1,1, \dots,1]^{\top} \in \mathbb{R}^{N}$,  and $\bm{B}=f_s(\bm{W}^{\top}\bm{X})\in\{0,1\}^{M\times N}$.\footnote{$f_s(\cdot)$ can be generalized to vector/matrix input, where an element-wise operation is applied.} $\bm{U} = [\bm{u},\dots,\bm{u}] \in \mathbb{R}^{M\times N}$, which  repeats $\bm{u} = [\mu_1,\dots,\mu_M]^{\top} \in \mathbb{R}^{M}$ $N$ times. 

Since the objective function depends on both the binary codes $\bm{B}$ and the projection matrix $\bm{W}$, we adopt an alternating optimization strategy. Specifically, $\bm{B}$ is updated with $\bm{W}$ fixed, and vice versa, until convergence.

\noindent\textbf{1) Update $\bm{B}$ while Fixing $\bm{W}$}: \quad 
Following the definition in Eq.~\eqref{eq:bmn}, a straightforward solution is, 
\begin{align}
\label{eqn:updateB1}
\bm{B} = f_s(\bm{W}^{\top}\bm{X}).
\end{align}

\noindent\textbf{2) Update $\bm{W}$ while Fixing $\bm{B}$}: \quad 
$f_s(\cdot)$ is a discrete nonlinear function, making $L$ non-differentiable w.r.t. $\bm{W}$. Following the strategy in~\cite{lu2015learning}, $\bm{B}$ is approximated by $\bm{W}^{\top}\bm{X}$ in Eq.~\eqref{eqn:objective2}, except the $L_1$ term. 
Then $L_1$ can be rewritten as, 
\begin{equation}
\begin{aligned}
\label{eqn:L1_2}
L_1 
    &={\Vert \bm{B}- \bm{W}^{\top} \bm{X} \Vert}^{2}\\
    &=\text{tr}(\bm{W}^{\top}\bm{X}\bm{X}^{\top} \bm{W})
    - 2\text{tr}(\bm{B}\bm{X}^{\top} \bm{W}) + \text{tr}(\bm{B}^{\top}\bm{B}),
\end{aligned}
\end{equation}
where $\text{tr}(\cdot)$ denotes the trace function. 
$L_2$ can be rewritten as,
\begin{equation}
\begin{aligned}
\label{eqn:L3_2}
L_2 &= {\Vert (\bm{B}-0.5)\bm{1}_N \Vert}^{2}   \\  
& = \text{tr}(\bm{W}^{\top}\bm{X} \bm{1}_N \bm{1}_N^{\top} \bm{X}^{\top}\bm{W}) \\
 &- N\text{tr}(\bm{1}_M^{\top}\bm{W}^{\top}\bm{X} \bm{1}_N)
 + 0.25MN^2, 
\end{aligned}
\end{equation}
where 
$\bm{1}_M = [1,1, \dots,1]^{\top} \in \mathbb{R}^{M}$. 
Denote $\overline{\bm{x}} = \frac{1}{N}\bm{X}\bm{1}_N \in \mathbb{R}^{D}$ as the mean PDV, and $\bm{M}= [\overline{\bm{x}},...,\overline{\bm{x}}]^{\top} \in \mathbb{R}^{D\times N}$. As $\bm{u}= \frac{1}{N}\bm{B}\bm{1}_N$, it is easy to show that $\bm{U} = \bm{W}^{\top}\bm{M}$ as $\bm{B}$ is approximated by $\bm{W}^{\top}\bm{X}$. Then, $L_3$ can be rewritten as, 
\begin{equation}
\begin{aligned}
\label{eqn:L3_2}
L_3 &=-\Vert\bm{B}-\bm{U}\Vert^2 \\
&=-\text{tr}\left(\bm{W}^{\top}(\bm{X}-\bm{M})(\bm{X}-\bm{M})^{\top}\bm{W}\right). 
\end{aligned} 
\end{equation}
$L_4$ can be rewritten as,
\begin{equation}
\begin{aligned}
\label{eqn:L1_2}
 L_4 
    &=\Vert{\bm{B}-\bm{B}\bm{A}}\Vert^2\\
    &=\text{tr}\left(\bm{W}^{\top}\left(\bm{X}(\bm{I}-\bm{A})(\bm{I}-\bm{A})^{\top}\bm{X}^{\top}\right)\bm{W}\right),
\end{aligned}
\end{equation}
where $\bm{I}\in\{0,1\}^{N\times N}$ is an identity matrix. 
Finally, 
\begin{equation}
\begin{aligned}
\label{eqn:objective3}
    L &= \text{tr}(\bm{W}^{\top}\bm{Q}\bm{W})-2\lambda_1 \text{tr}({\bm{B}\bm{X}^{\top}\bm{W}}) + \lambda_1\text{tr}(\bm{B}^{\top}\bm{B}) \\
    &\quad - \lambda_2 N \text{tr}(\bm{1}_M^{\top}\bm{W}^{\top}\bm{X}\bm{1}_N) + 0.25MN^2, 
\end{aligned}
\end{equation}
where
\begin{equation}
\begin{aligned}
\label{eqn:Q}
\bm {Q}&\;\triangleq\; \bm{X}(\bm{I}-\bm{A})(\bm{I}-\bm{A})^{\top}\bm{X}^{\top} + \lambda_1\bm{X}\bm{X}^{\top} \\
&+\lambda_2\bm{X}\bm{1}_N\bm{1}_N^{\top}\bm {X}^{\top} - \lambda_3 (\bm{X}-\bm{M})(\bm{X}-\bm {M})^{\top}.
\end{aligned}
\end{equation}
After discarding the terms that do not contain $\bm{W}$, the optimal $\bm{W}$ can be obtained as follows, 
\begin{equation}
\begin{aligned}
\label{eqn:updateW}
\mathop{\min} _{\bm {W}}L(\bm {W})
&= \text{tr}(\bm{W}^{\top}\bm {Q}\bm{W}) - 2 \lambda_1\text{tr}(\bm{B}\bm{X}^{\top}\bm {W}) \\ 
& \quad- \lambda_2 N \text{tr}(\bm{1}_M^{\top}\bm{W}^{\top}\bm{X}\bm{1}_N), \\ 
& s.t. \quad \bm{W}^{\top}\bm{W}=\bm{I}. 
\end{aligned}
\end{equation}

The orthogonality constraint \( \mathbf{W}^\top\mathbf{W} = \mathbf{I} \) in Eq.~\eqref{eqn:updateW} makes direct Euclidean optimization unsuitable. We therefore recast the problem as an unconstrained optimization on the Stiefel manifold \(\mathcal{S}(D, M)\), whose points naturally satisfy orthogonality. Then, we solve this Riemannian problem via gradient descent adapted to the manifold geometry. The Euclidean gradient \( \frac{\partial L}{\partial\bm{W}} \) is projected onto the tangent space at \(\bm{W}\) to obtain the Riemannian gradient. This direction is then used in a curvilinear search that preserves orthogonality via retraction. The update employs the Sherman-Morrison-Woodbury formula for efficient inversion~\cite{wang2024new}, with step size selected by the Armijo-Wolfe rule to ensure convergence~\cite{han2023riemannian}. 
The algorithm alternately updates the projection matrix $\bm{W}$ on the Stiefel manifold and the binary codes $\bm{B}$, as detailed in Algo.~\ref{alg:W}.

\begin{algorithm} [!t]
\caption{Procedures for PDV hashing. 
}
\label{alg:W}
\textbf{Input:} PDV matrix $\bm{X} \in\mathbb{R}^{D\times N}$, convergence parameter $\theta$. \\ 
\textbf{Output:} Optimal $\bm{W}\in\mathbb{R}^{D\times M}$.
\begin{algorithmic}[1]
\STATE Initialize $\bm{W}$ as the top $M$ eigenvectors of $\bm{X}\bm{X}^{\top}$ corresponding to the largest eigenvalues. 
\STATE Derive the neighbourhood matrix $\{\Tilde{\bm{a}}_{n}\}_{n=1}^N$ as in Eq.~(\ref{eqn:LLE_Lagrangian5}).\\
\REPEAT
\STATE Update $\bm{B}$ while fixing $\bm{W}$ as in Eq.~(\ref{eqn:updateB1}).
\STATE Calculate $L_1$, $L_2$, $L_3$ and $L_4$ as in Eqs.~\eqref{eqn:quantiztion_loss}, \eqref{eqn:max_info},
\eqref{eqn:L4} and \eqref{eqn:J1}, respectively. 
\STATE Calculate the final loss function $L$ as in Eq.~\eqref{eqn:objective}. 
\STATE Rewrite $L$ in a matrix form as in Eq.~\eqref{eqn:updateW}. 
\STATE Derive the gradient $\frac{\partial L}{\partial\bm{W}}$ and update $\bm{W}$ using the curvilinear search algorithm with Sherman-Morrison-Woodbury formula~\cite{wang2024new}.
\UNTIL $|\frac{\partial L}{\partial\bm{W}}|<\theta$.
\RETURN Optimal projection matrix $\bm{W}$. 
\end{algorithmic}  
\end{algorithm}

We now analyze the computational complexity of Algo.~\ref{alg:W}. The main steps include: 
1) truncated SVD initialization of $\bm{W}$ in
$O(NDM)$;
2) neighbourhood computation via KD-tree and KNN in $O(N \log N + N(DK^2 + K^3))$; 
3) binary code update in $O(NDM)$;
4) precomputation of $\bm{Q}$ in $O(NDK + ND^{2})$;
and 5) $T$ rounds of loss evaluation and manifold-based $\bm{W}$ update, each costing $O(NDM + NM^{2})$ under $D$, $M \ll N$.  
The total complexity is $O\big(N \log N + N\big(\big(DM+D^2+DK^2 +K^3) + T(DM + M^{2})\big)\big).$ 
The complexity is dominated by \(O(N \log N)\), offering good scalability for practical dynamic texture recognition tasks. For truly massive-scale applications, approximate nearest-neighbor methods~\cite{malkov2020efficient} can be leveraged to further improve computational efficiency.


\subsection{Dictionary Learning for Binary Vectors}
Although PDV hashing yields discriminative binary codes, residual redundancy may still exist. To further enhance compactness and robustness, we apply dictionary learning to the binary codes. Specifically, k-means clustering is used to learn a dictionary of $C$ codewords, and each binary code is mapped to its nearest codeword. The resulting histogram of codewords serves as the final feature representation.

To capture texture information at multiple spatial and temporal scales, we extend the multi-scale strategy described in~\cite{bi20212dlcolbp}. PDVs are first extracted from cubic neighbourhoods of varying sizes, such as \(P=3\) and \(P=5\), which correspond to neighbourhoods of 26 and 124 points, respectively. For each scale, a dedicated set of hash functions is learned to map the PDVs into binary codes. The resulting multi-scale binary features are then concatenated and further compressed via PCA to reduce dimensionality while preserving discriminative information. This multi-scale design allows LP\textsuperscript{2}DH to integrate discriminant patterns across different granularities, yielding a compact and highly discriminative feature representation.

\section{Experimental Results}

\subsection{Experimental Settings}
\subsubsection{Compared Methods} 
Our LP\textsuperscript{2}DH is compared with representative methods from six main categories of DT recognition. The selected baselines cover a wide spectrum of established approaches, ensuring a comprehensive comparison.
\textbf{1)~Optical-flow-based methods} are included as they explicitly model motion, a fundamental characteristic of dynamic textures. We compare with FDT~\cite{nguyen2018directional} and its extension FD-MAP~\cite{nguyen2018directional}, which are seminal works in capturing and quantizing motion patterns. 
\textbf{2)~Model-based methods} offer a generative perspective. We select HEM-DTM~\cite{mumtaz2012clustering}, JVDL~\cite{wei2017dynamical}, and MixSHMM~\cite{xing2023dynamic} for their diverse modelling strategies using LDS clustering, dictionary learning, and HMM mixtures, respectively.
\textbf{3)~Geometry-based methods} provide scale-invariant analysis. We choose DFS~\cite{xu2015classifying} and STLS~\cite{quan2017spatiotemporal} for their robustness to viewpoint and illumination changes via fractal and lacunarity analysis. 
\textbf{4)~Filter-based methods} are known for their robustness to noise. Our comparison includes MBSIF-TOP~\cite{arashloo2014dynamic}, DNG\textsubscript{P}~\cite{rivera2015spatiotemporal}, B3DF\_SMC~\cite{zhao2019dynamic}, HoGF~\cite{nguyen2021prominent}, and SOE~\cite{derpanis2012dynamic}, which represent key advancements in spatiotemporal filtering.
\textbf{5)~Local-feature-based methods}, particularly LBP variants, are classical and widely used for their efficiency. We evaluate popular LBP descriptors (VLBP~\cite{zhao2007dynamic}, LBP-TOP~\cite{zhao2007dynamic}) and more recent variants (CVLBC~\cite{zhao2018dynamic}, DDLBP~\cite{ren2015learning}, DBRF~\cite{zhao2023dynamic}, RUBIG~\cite{nguyen2020rubik}, MEMDP~\cite{nguyen2020momental}) to cover the evolution in this category.
\textbf{6)~Learning-based methods} represent the state-of-the-art. We compare both traditional learning on handcrafted features (\eg, OTDL~\cite{quan2015dynamic}, EKDL~\cite{quan2016equiangular}, PI-LBP~\cite{ren2013dynamic}) and deep learning models, from early networks (AlexNet/DT-GoogleNet~\cite{andrearczyk2018convolutional}) to modern self-supervised models (Unsupervised-CNN~\cite{wang2015unsupervised}, ClipOrder~\cite{misra2016shuffle}, Geometry~\cite{gan2018geometry}, STS~\cite{wang2022self})  and 3D architectures (SOE-Net~\cite{hadji2017spatiotemporal}).

\subsubsection{Parameter settings}
We follow the standard evaluation protocol of each dataset and report classification accuracy throughout. Comparative results for baselines are taken from the published literature. 
To directly assess descriptor discriminability, we use a nearest-neighbour (NN) classifier with cosine distance for LP\textsuperscript{2}DH. The dictionary size is set to $C=3000$ on all datasets; the binary code length is $M = 16$ for $P = 3$ and $M = 40$ for $P = 5$; and the number of nearest neighbours in LLE is $K = 10$. The hyperparameters in Eq.~\eqref{eqn:objective3} are set to $\lambda_1 = 10^1$, $\lambda_2 = 1$, and $\lambda_3 = 10^3$, with sensitivity analysis reported in Sec.~\ref{sec:Parameter}. 

\begin{table*}[!t]
\caption{Comparisons to state-of-the-art methods in terms of recognition rates in \%  on the UCLA and DynTex++ dataset.}
\label{ucla_table}
\centering
\begin{threeparttable}
\begin{tabular}{|l|l| l |l| l |l|}
\hline
\multirow{2}{*}{\textbf{Class}} & \multirow{2}{*}{\textbf{Method}} & \multicolumn{3}{c|}{\textbf{UCLA}} & \multirow{2}{*}{\textbf{DynTex++}} \\ \cline{3-5}
\multirow{2}{*}{~} & \multirow{2}{*}{~} & 50-class  & 9-class  & Average & \multirow{2}{*}{~} \\
\hline
\multirow{2}{*}{Optical-flow-based}  
                    & {FDT\dag}~\cite{nguyen2018directional}    & 99.00 & 97.70 & 98.35& 95.31 \\
                    & FD-MAP~\cite{nguyen2018directional} & 99.00 & 99.35 & 99.18 & 95.69 \\
\hline
\multirow{3}{*}{Model-based}         
                    & {HEM-DTM}~\cite{mumtaz2012clustering} & 96.55 & - & - & - \\
                    & JVDL~\cite{wei2017dynamical} & 97.83 & - & - & 91.8 \\
                    & {MixSHMM\dag}~\cite{xing2023dynamic} & $^{*}$\textbf{100} & $^{*}$\textbf{99.60} & $^{*}$\textbf{99.80} & 95.37 \\
\hline
\multirow{2}{*}{Geometry-based}      
                    & DFS~\cite{xu2015classifying} & \textbf{100} & 97.50  & 98.75 & 91.70 \\ 
                    & STLS~\cite{quan2017spatiotemporal}  & $^{*}$99.50 & $^{*}$97.40 & $^{*}$98.45 & 94.50 \\
\hline
\multirow{5}{*}{Filter-based}	    
                    & {MBSIF-TOP\dag}~\cite{arashloo2014dynamic}& 99.50 & 98.75$^N$ & 99.13$^N$ & 97.17$^N$ \\
                    & DNG\textsubscript{P}~\cite{rivera2015spatiotemporal}& - & \textbf{99.60} & - & 93.80 \\
                    & B3DF\_SMC~\cite{zhao2019dynamic}& 99.50$^N$ & 98.85$^N$ & 99.18$^N$ & 95.58$^N$ \\
                    & HOGF\textsuperscript{2D}~\cite{nguyen2021prominent} &$^{*}$\textbf{100}& $^{*}$99.20 & \underline{$^{*}$99.60} & 97.19 \\
                    & HOGF\textsuperscript{3D}~\cite{nguyen2021prominent} &$^{*}$\textbf{100}& \underline{$^{*}$99.55} & $^{*}$99.78 & 97.63 \\
\hline
\multirow{7}{*}{Local-feature-based} 
                    & {VLBP}~\cite{zhao2007dynamic} & 89.50$^N$ & 96.30$^N$ & 92.90$^N$ & 94.98$^N$ \\
                    & {LBP-TOP}~\cite{zhao2007dynamic} & 94.50$^N$ & 96.00$^N$  & 95.25$^N$ & 94.05$^N$ \\
                    & {CVLBC}~\cite{zhao2018dynamic} & 99.00$^N$ & 99.20$^N$ & 99.10$^N$ & 91.31$^N$ \\
                    & DDLBP$\dag$~\cite{ren2015learning} & \textbf{100} & 98.40 & 99.20 & 95.90 \\
                    & RUBIG~\cite{nguyen2020rubik} & $^{*}$\textbf{100}& $^{*}$99.20 & \underline{$^{*}$99.60} & 97.08 \\
                    & MEMDP~\cite{nguyen2020momental} & \textbf{100}& 98.90 & 99.45 & 96.03 \\
                    & {DBRF\dag}~\cite{zhao2023dynamic} & \textbf{100$^N$} & 99.00$^N$ & 99.50$^N$ & 95.18$^N$ \\
\hline
\multirow{10}{*}{Learning-based}      
                    & {DL-PEGASOS}~\cite{ghanem2010maximum} & 97.50 & 95.60 & 96.55 & 63.70  \\
                    & {PI-LBP}~\cite{ren2013dynamic} & \textbf{100$^N$} & 98.20$^N$ & 99.10 & 92.40$^N$ \\
                    & OTDL~\cite{quan2015dynamic} & \underline{99.80} & 98.20 & 99.00 & 94.70 \\
                    & EKDL~\cite{quan2016equiangular} & - & - & - & 93.40 \\
                    & RNNs~\cite{junior2019randomized} &97.05$^N$ &98.54$^N$ & 97.80$^N$ &96.51$^N$ \\ 
                    & SOE-Net~\cite{hadji2017spatiotemporal} & - & - & - & 94.40$^D$ \\
                    & AlexNet~\cite{andrearczyk2018convolutional} & 99.50$^D$ & 98.05$^D$ & 98.78$^D$ & 98.18$^D$ \\
                    & DT-GoogleNet~\cite{andrearczyk2018convolutional} & 99.50$^D$ & 98.35$^D$ & 98.93$^D$ & \textbf{98.58}$^D$ \\\cline{2-6}
                    & \multirow{2}{*}{\textbf{Proposed LP\textsuperscript{2}DH}} & \textbf{100}$^N$ & 99.30$^N$  & 99.65$^N$ & \multirow{2}{*}{\underline{98.52$^N$}} \\
                    & \multirow{2}{*}{~} & $^{*}$\textbf{100}$^{N}$ & $^{*}$\textbf{99.60}$^{N}$ & $^{*}$\textbf{99.80}$^N$ & \multirow{2}{*}{~} \\
\hline
\end{tabular}

    \begin{tablenotes}
        \footnotesize
        \item[$\dag$] Results are sourced from the original publications; otherwise, they are taken from~\cite{nguyen2021comprehensive}. 
        \item[$N$] denotes results using an NN classifier; $^D$ denotes results from a deep learning model. Unmarked entries report results using an SVM, as per the original papers.                  
        \item[*] The video size is $160\times 110\times 75$.

    \end{tablenotes}
\end{threeparttable}
\end{table*}

\subsection{Comparison Results on UCLA Dataset}
The UCLA dataset~\cite{doretto2003dynamic} contains 200 DT videos from 50 scenes (4 videos per scene). Each video has 75 frames of size 160 × 110. Most prior work uses cropped videos of 48 × 48 × 75 (default size), while some methods report results on the original resolution. Following the common evaluation protocols in~\cite{nguyen2021prominent, zhao2023dynamic}, we report results under two class settings. 
In the 50-class setting, we perform standard 4-fold cross validation~\cite{nguyen2021prominent} and report the average over four folds. In the 9-class setting~\cite{nguyen2021prominent}, the 50 scenes are merged into nine categories, and we randomly select half of the videos in each category for training and the remainder for testing, repeating the process for 20 trials and reporting the average accuracy.

Tab.~\ref{ucla_table} summarizes the results on the UCLA dataset. Three key observations emerge from the results.  
1)~On this relatively small dataset, the proposed method achieves the highest average accuracy of 99.80\%, although many approaches show near-perfect performance. This saturation effect arises from the dataset's limited intra-class variability and static backgrounds. Our PDV-hashing module counteracts this by regularizing the feature space: its bit-balance and variance-maximization constraints reduce quantization loss and preserve local structures, which helps mitigate overfitting and improves generalization on small-sample data.
2)~LP\textsuperscript{2}DH attains superior accuracy using only a nearest-neighbour classifier, while most competing methods rely on more complex classifiers (\eg, SVMs or deep networks). Compared to DBRF~\cite{zhao2023dynamic}, which also employs an NN classifier, our method raises the average accuracy from 99.50\% to 99.80\%. This confirms that the performance gain derives from the high-quality feature encoding produced by the hashing and dictionary-learning pipeline, not from classifier sophistication.
3)~LP\textsuperscript{2}DH also outperforms the leading deep learning model, DT-GoogleNet~\cite{andrearczyk2018convolutional} (98.93\% vs. 99.80\%). This observed discrepancy aligns with a recognized challenge of deep networks: their high model complexity can render them prone to overfitting on datasets of limited size. In contrast, our proposed pipeline, based on hashing and dictionary learning, is inherently parameter-efficient. This design facilitates robust generalization even with scarce training data, while preserving strong discriminative performance.

\begin{figure}[!t] 
	\centering
        \includegraphics[width=0.8\linewidth]{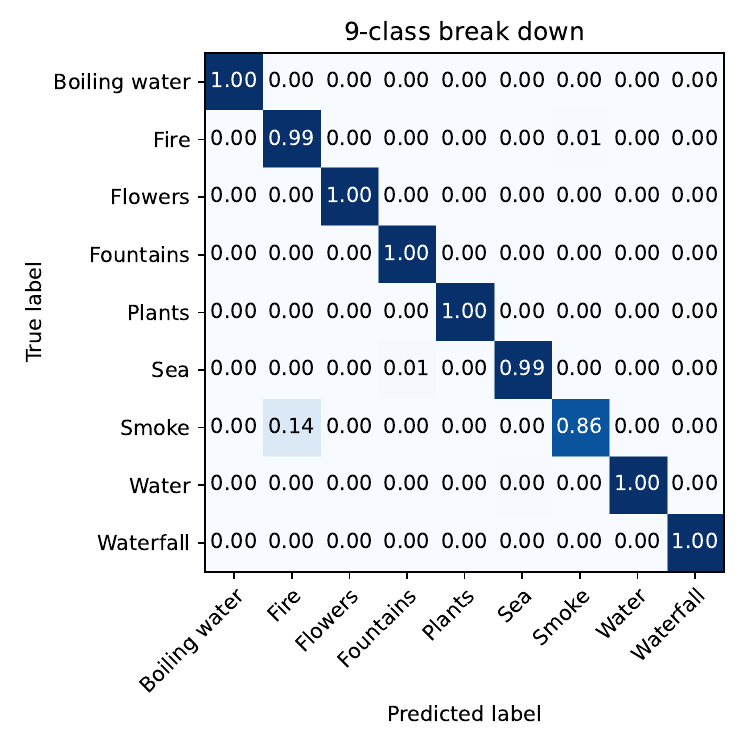}
	\caption{Confusion matrix for the 9-class setting on UCLA~\cite{doretto2003dynamic}.}
	\label{fig:confusion} 
\end{figure}

Fig.~\ref{fig:confusion} shows the confusion matrix for the 9-class setting on UCLA~\cite{doretto2003dynamic}. A small amount of confusion appears between ``smoke'' and ``fire'', which is expected given their visual and dynamic similarity (\eg, partially occluded flames or smoke-dominant frames that share local texture and motion cues). This error pattern suggests that LP\textsuperscript{2}DH focuses on low-level dynamics and may occasionally struggle when semantic classes share highly overlapping spatiotemporal signatures.

\subsection{Comparison Results on DynTex++ Dataset} 
DynTex++~\cite{ghanem2010maximum} is a large benchmark dataset for DT recognition. It contains 36 classes of dynamic textures, each with 100 videos of size $50\times50\times50$. 
Following the standard protocol in~\cite{arashloo2014dynamic, zhao2023dynamic}, we randomly select half of the videos in each class for training and the remainder for testing, repeat the split five times, and report the average accuracy in Tab.~\ref{ucla_table}.

On the DynTex++ dataset, LP\textsuperscript{2}DH achieves 98.52\% accuracy, closely matching the deep learning model DT-GoogleNet~\cite{andrearczyk2018convolutional} (98.58\%) and outperforming all other methods. The 0.89\% gain over HOGF\textsuperscript{3D}~\cite{nguyen2021prominent} stems from a fundamental difference in feature encoding strategy. HOGF\textsuperscript{3D} uses gradient filtering, a low-pass operation that can obscure discriminative high-frequency texture details. In contrast, LP\textsuperscript{2}DH utilizes learning-based hashing to explicitly preserve structural relationships within local pixel differences, enforced through bit-balance, variance, and locality constraints. Together with dictionary-based semantic aggregation, LP\textsuperscript{2}DH captures subtle inter-class variations that are often lost in conventional handcrafted pipelines.

\begin{figure}[!t] 
	\centering
 \includegraphics[width=1\linewidth]{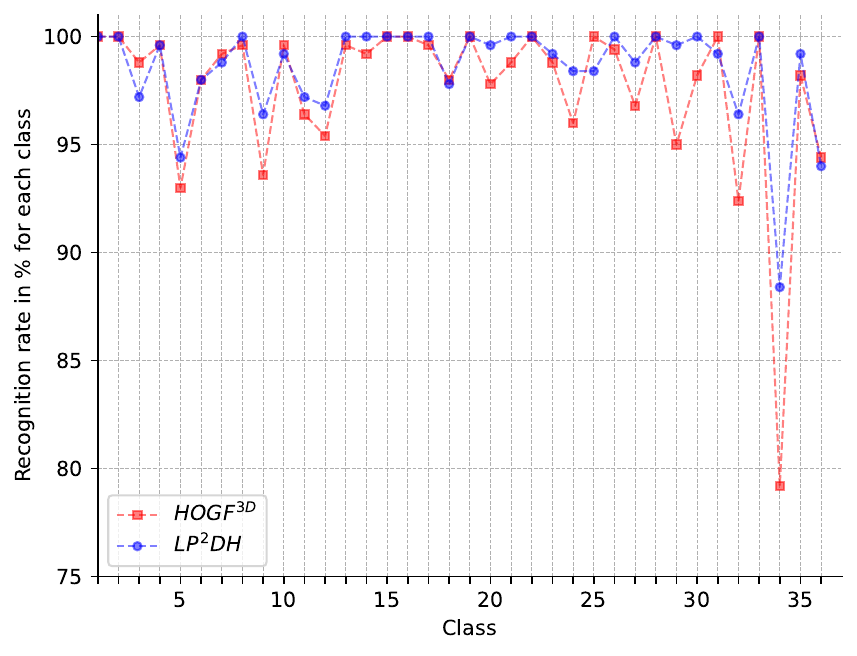}
 \caption{The per-class recognition rates of HOGF\textsuperscript{3D}~\cite{nguyen2021prominent} and LP\textsuperscript{2}DH on the DynTex++ dataset~\cite{ghanem2010maximum}.} 
\label{fig:dyntex_compare} 
\end{figure}

Fig.~\ref{fig:dyntex_compare} shows that LP\textsuperscript{2}DH  outperforms HOGF\textsuperscript{3D}~\cite{nguyen2021prominent}  in nearly all classes, with gains above 10\% on several challenging categories. The largest improvements occur for fine-grained motions or subtle intensity changes (\eg, flowing water, shimmering leaves, and swirling smoke), where gradient-based operators may miss discriminative temporal cues. By preserving PDV neighbourhood structure through locality-sensitive hashing, LP\textsuperscript{2}DH retains these cues and better separates visually similar yet dynamically distinct textures.

\subsection{Comparison Results on YUPENN Dataset}
The YUPENN dataset~\cite{derpanis2012dynamic} contains 420 videos from 14 classes (30 videos per class). 
We compare LP\textsuperscript{2}DH with handcrafted feature-based learning methods (\eg, SOE~\cite{derpanis2012dynamic} and SFA~\cite{theriault2013dynamic}), and with deep-learning models (\eg, unsupervised-CNN~\cite{wang2015unsupervised}, ClipOrder~\cite{misra2016shuffle}, Geometry~\cite{gan2018geometry} and STS~\cite{wang2022self}). 
For STS~\cite{wang2022self}, we report results with three backbones: 3D-ResNet18 (R3D-18), R(2+1)D that factorizes 3D convolution into 2D spatial and 1D temporal operations, and C3D with five 3D convolutional blocks. We follow the leave-one-out protocol in~\cite{wang2022self}: in each fold, one video per class is used for testing and the remaining videos are used for training.

The experimental results on the YUPENN dataset (Tab.~\ref{YUP_table}), with comparison data sourced from~\cite{wang2022self}, reveal three key findings. 
1)~LP\textsuperscript{2}DH achieves state-of-the-art performance on the YUPENN dataset, surpassing both deep and handcrafted methods. This gain comes from two key design principles. First, locality-preserving hashing maintains neighbourhood topology to capture fine motion patterns. Second, constrained joint optimization (via bit balance and variance maximization) aligns binary codes with the statistical properties of dynamic textures. 
2)~LP\textsuperscript{2}DH outperforms the leading deep learning method, STS-C3D~\cite{wang2022self}, by 1.19\%. This shows that its combination of locality-preserving hashing and dictionary learning can serve as a data-efficient alternative to data-hungry deep models.
3)~LP\textsuperscript{2}DH exceeds conventional handcrafted methods such as SOE~\cite{derpanis2012dynamic} and SFA~\cite{theriault2013dynamic} by over 10\%. This gain demonstrates that iterative optimization of hashing functions and visual dictionaries enables the model to learn discriminative patterns beyond static feature engineering. 
These results demonstrate that well-designed, lightweight models with domain-informed geometric priors can surpass complex deep architectures in dynamic texture recognition. 

\begin{table}[!t]
\caption{Comparison on the YUPENN dataset. 
$^N$ marks nearest-neighbour classifier; $^D$ marks deep learning models.}
\label{YUP_table}
\centering
\begin{threeparttable}
\begin{tabular}{l l}
\toprule
\textbf{Method} & \textbf{Accuracy (\%) }\\
\midrule
SOE~\cite{derpanis2012dynamic} & 80.70$^N$ \\
SFA~\cite{theriault2013dynamic}& 85.50$^N$ \\
Unsupervised-CNN~\cite{wang2015unsupervised}& 70.50$^D$ \\
ClipOrder~\cite{misra2016shuffle} & 76.70$^D$ \\
Geometry~\cite{gan2018geometry} & 86.90$^D$ \\ 
STS, 3D-ResNet18~\cite{wang2022self}  & 92.90$^D$ \\
STS, R(2+1)D~\cite{wang2022self}  & 94.30$^D$ \\
STS, C3D~\cite{wang2022self}  & \underline{95.00}$^D$ \\ \midrule
\textbf{LP\textsuperscript{2}DH}&\textbf{96.19}$^N$ \\
\bottomrule
\end{tabular}
\end{threeparttable}
\end{table}

\subsection{Discussions of Results across Three Datasets}
The evaluation across three datasets reveals several key insights.
1)~LP\textsuperscript{2}DH delivers consistently strong performance across all benchmarks. It achieves top accuracy on UCLA under both class settings, ranks second on DynTex++, and outperforms all competitors on YUPENN. This cross-dataset robustness confirms that our method is not tailored to any single dataset, but offers a broadly effective framework for DT recognition.
2)~DT-GoogleNet marginally outperforms LP\textsuperscript{2}DH on DynTex++ (98.58\% vs. 98.52\%), but requires large-scale training data and performs poorly on smaller datasets like UCLA. In contrast, LP\textsuperscript{2}DH performs well on both large and small datasets, showing strong data efficiency. Its PDV-based design learns a compact, discriminative representation, avoiding deep models' data hunger without losing accuracy. 
3) On YUPENN, LP\textsuperscript{2}DH outperforms recent deep models like STS~\cite{wang2022self}, showing that a well-designed feature-learning pipeline can outperform end-to-end deep architectures. By combining hashing-based embedding with a visual dictionary, it produces texture-specific features that reduce overfitting and improve generalization under data constraints.

\subsection{Ablation Studies}
\subsubsection{Ablation of Key Components}
We evaluate two core components of LP\textsuperscript{2}DH: the PDV hashing module and the locality-preserving criterion, by incrementally adding them to the VLBP baseline (Tab.~\ref{ablation}). The PDV hashing module provides clear gains over VLBP, including +9.50\% and +1.80\% on the two UCLA protocols, +2.53\% on DynTex++, and above +13\% on YUPENN, indicating that raw pixel differences alone are insufficient. By projecting high-dimensional PDVs into a compact binary space, our hashing step reduces noise and keeps key discriminative patterns for multi-class recognition. Adding the locality-preserving criterion boosts accuracy by 1.00–1.50\% across datasets, showing the value of preserving PDV neighbourhood structure. This geometric constraint yields compact, semantically structured codes that are both more generalizable and geometrically meaningful.

\begin{table}[!t]
\caption{Ablation study of major components, PDV hashing framework and locality-preserving criterion.
These components are gradually added on top of the baseline model VLBP.} 
\label{ablation}
\centering
\begin{tabular}{l|c|c|c|c}
\hline
\multirow{2}{*}{Components} & \multicolumn{2}{c|}{\textbf{UCLA}} & \multirow{2}{*}{DynTex++} & \multirow{2}{*}{YUPENN} \\ \cline{2-3}
\multirow{2}{*}{~} & 50-class & 9-class & \multirow{2}{*}{~}\\
\hline
VLBP~\cite{zhao2007dynamic}     &89.50 &96.30  & 94.98 & 81.70 \\
+ Hashing framework         &99.00 &98.10  & 97.51 & 94.76 \\
+ Locality preserving      &100 &99.60 & 98.52 & 96.19 \\
\hline
\end{tabular}
\end{table}

\begin{figure*}[htbp] 
	\centering
	\includegraphics[width=1\linewidth]{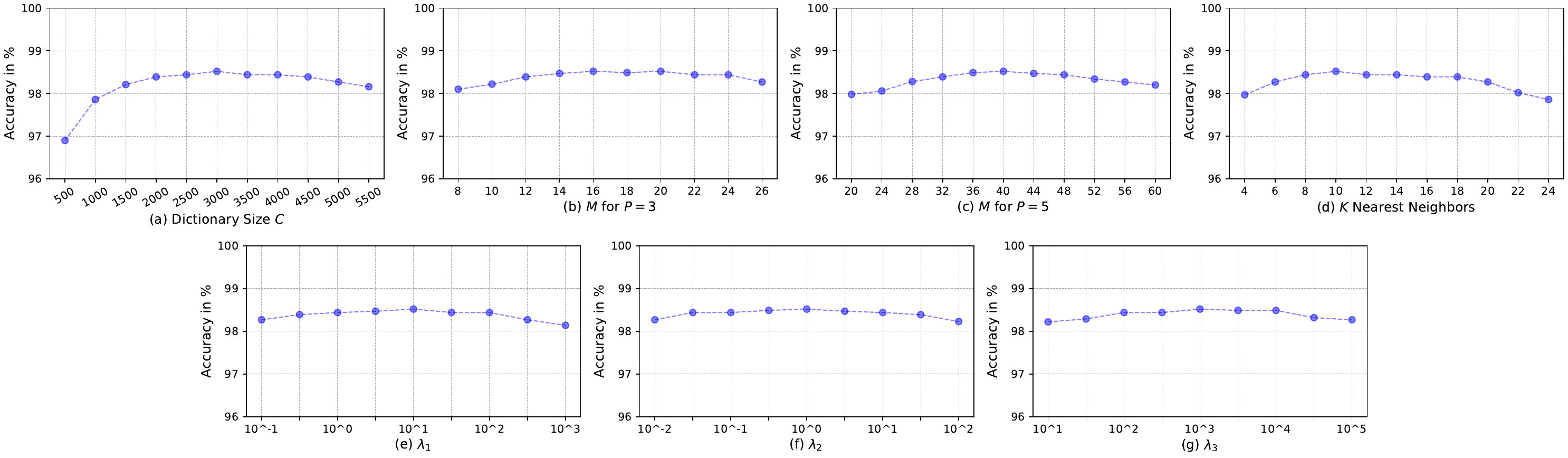}
	\caption{Ablation of LP\textsuperscript{2}DH with different (a) dictionary size, (b) binary code length for $P=3$, (c) binary code length for $P=5$, (d) the number of nearest neighbours, (e) $\lambda_1$, (f) $\lambda_2$, and (g) $\lambda_3$.} 
	\label{fig:lineChart} 
\end{figure*}

\subsubsection{Ablation of Key Parameters}
\label{sec:Parameter}
Our parameter sensitivity analysis (Fig.~\ref{fig:lineChart}) explores how key parameters affect LP\textsuperscript{2}DH’s performance and robustness. Each parameter governs a specific learning aspect, and its optimal setting reflects a careful trade-off in representation quality.

\noindent\textbf{Dictionary size} $C$ controls representational capacity. A small $C$ (\eg, 500) increases quantization error and loses detail, while the optimal $C=3000$ balances capacity and generalization. Larger dictionaries memorize dataset-specific noise rather than semantic patterns, indicating that dictionary complexity should match the data’s intrinsic dimensionality.

\noindent\textbf{Code length} $M$ balances information retention and stability. Short codes ($M<16$ for $P=3$) discard critical texture cues. Optimal values (16 for $P=3$, 40 for $P=5$) preserve entropy and resist noise. Excessively long codes amplify Hamming distances between similar patterns, reducing invariance. Lastly, as $P$ grows, code length $M$ must also grow to encode greater spatial complexity while preserving stability. 

\noindent\textbf{Nearest-neighbour size} $K$ affects manifold approximation. Too few neighbours ($K<10$) yield unstable local estimates and overfit noise. Too many ($K>10$) blur geometry by violating local linearity. The optimal $K=10$ maintains a faithful and consistent approximation across the dataset.

\noindent\textbf{Regularization weights} $\lambda_1$, $\lambda_2$, and $\lambda_3$ govern objective balance. Peak performance occurs at $\lambda_1=10^1$, $\lambda_2=1$, $\lambda_3=10^3$. Smooth performance decay around these values shows a well-behaved objective and confirms LP\textsuperscript{2}DH’s robustness to reasonable parameter changes. 

Overall, these ablation results demonstrate that LP\textsuperscript{2}DH's architecture is both logically grounded and empirically stable, supporting its effectiveness across diverse settings. 

\subsection{Runtime Comparison}
A comparative runtime analysis (Tab.~\ref{tab:runtime_dyntexpp}) demonstrates the superior computational efficiency of LP\textsuperscript{2}DH. All experiments were performed on a workstation equipped with dual NVIDIA RTX A5000 GPUs and an Intel Xeon Gold 5218 CPU, utilizing author-provided implementations. For DT-GoogleNet~\cite{andrearczyk2018convolutional}, each video is processed by extracting 50 slices per orthogonal plane, resulting in a total of $1,800 \times 3 \times 50 = 270,000$ slices for the complete test set. DT-GoogleNet employs an adapted Inception-v1 architecture, configured for $50\times50$-pixel inputs to match the dimensions of the extracted planar slices, with a batch size of 64~\cite{andrearczyk2018convolutional}. Reported test times are end-to-end measurements, covering the complete processing pipeline from video input to prediction output for all test videos.  

As shown in Tab.~\ref{tab:runtime_dyntexpp}, LP\textsuperscript{2}DH reduces training time to approximately 22\% of that required by DT-GoogleNet~\cite{andrearczyk2018convolutional}, albeit with a longer inference duration. It should be noted that DT-GoogleNet relies on GPU acceleration for parallel computation, whereas LP\textsuperscript{2}DH is executed entirely on the CPU.   
Among non-deep methods, LP\textsuperscript{2}DH achieves a superior speed‑accuracy trade‑off, outperforming contenders such as HOGF\textsuperscript{3D}~\cite{nguyen2021prominent} and RUBIG~\cite{nguyen2020rubik} in both metrics. Furthermore, it avoids the high computational costs of STLS~\cite{quan2017spatiotemporal} and FD-MAP~\cite{nguyen2018directional}. This efficiency, coupled with its CPU‑only execution, makes LP\textsuperscript{2}DH particularly suitable for resource‑constrained environments or systems requiring frequent retraining. 
\begin{table}[!t]
\centering
\caption{Runtime comparison on the DynTex++ dataset.}
\label{tab:runtime_dyntexpp}
\begin{tabular}{l|l|r|r}
\hline
Category & Method & Train (s) & Test (s) \\
\hline
Optical-flow-based   & FD-MAP~\cite{nguyen2018directional}            & 2,567 & 2,636 \\
Model-based          & MixSHMM~\cite{xing2023dynamic}                 & 3,234 & 2,467 \\
Geometry-based       & STLS~\cite{quan2017spatiotemporal}             & 13,206 & 13,109 \\
Filter-based         & HOGF\textsuperscript{3D}~\cite{nguyen2021prominent} & 1,898 & 1,904 \\
Local-feature-based  & RUBIG~\cite{nguyen2020rubik}                   & \textbf{1,589} & 1,597 \\\hline
\multirow{2}{*}{Learning-based}
        & DT-GoogleNet~\cite{andrearczyk2018convolutional}  & 9,033 & \textbf{129}  \\ \cline{2-4}
        & \textbf{LP$^{2}$DH}                                      & 2,038 & 915  \\
\hline
\end{tabular}
\end{table}

\section{Conclusion and Future Work}
\noindent\textbf{Theoretical Significance:} \quad
This paper introduces LP\textsuperscript{2}DH, a unified hashing framework for dynamic texture recognition. It jointly optimizes quantization accuracy, information entropy, discriminability, and locality preservation, extending spectral hashing with a manifold-inspired criterion. By performing optimization on the Stiefel manifold, LP\textsuperscript{2}DH preserves the topology of PDVs and improves convergence through geometrically constrained updates. Its two-stage hashing and dictionary learning strategy also enables effective dimensionality reduction while maintaining strong representational capacity.

\noindent\textbf{Practical Significance:} \quad
LP\textsuperscript{2}DH achieves state-of-the-art performance on benchmark datasets (\eg, 99.80\% on UCLA, 98.52\% on DynTex++, and 96.19\% on YUPENN), using only a nearest-neighbour classifier. Its compact features enable efficient deployment in resource-constrained applications. The method also requires little training data, making it well-suited for annotation-scarce scenarios. Ablation studies verify the contribution of each component, supporting reliable and reproducible real-world implementation.

\noindent\textbf{Limitations and Future Work:} \quad
Despite its strengths, LP\textsuperscript{2}DH has two main limitations. First, its multi-objective design involves inherent trade-offs, limiting universal optimality. Future work will employ Pareto front analysis~\cite{kalyanmoy2024identifying} to adaptively balance objectives across different applications.  
Second, the two-stage pipeline prevents end-to-end optimization: although designed for modularity and training stability, it isolates dictionary learning from hashing optimization. As a result, the projection matrix receives no direct feedback from classification or clustering objectives. A unified, end-to-end framework is hence an important direction for future work.

\bibliographystyle{IEEEtran}
\normalem
\bibliography{mybib}

\begin{IEEEbiography}[{\includegraphics[width=1in,height=1.25in,clip]{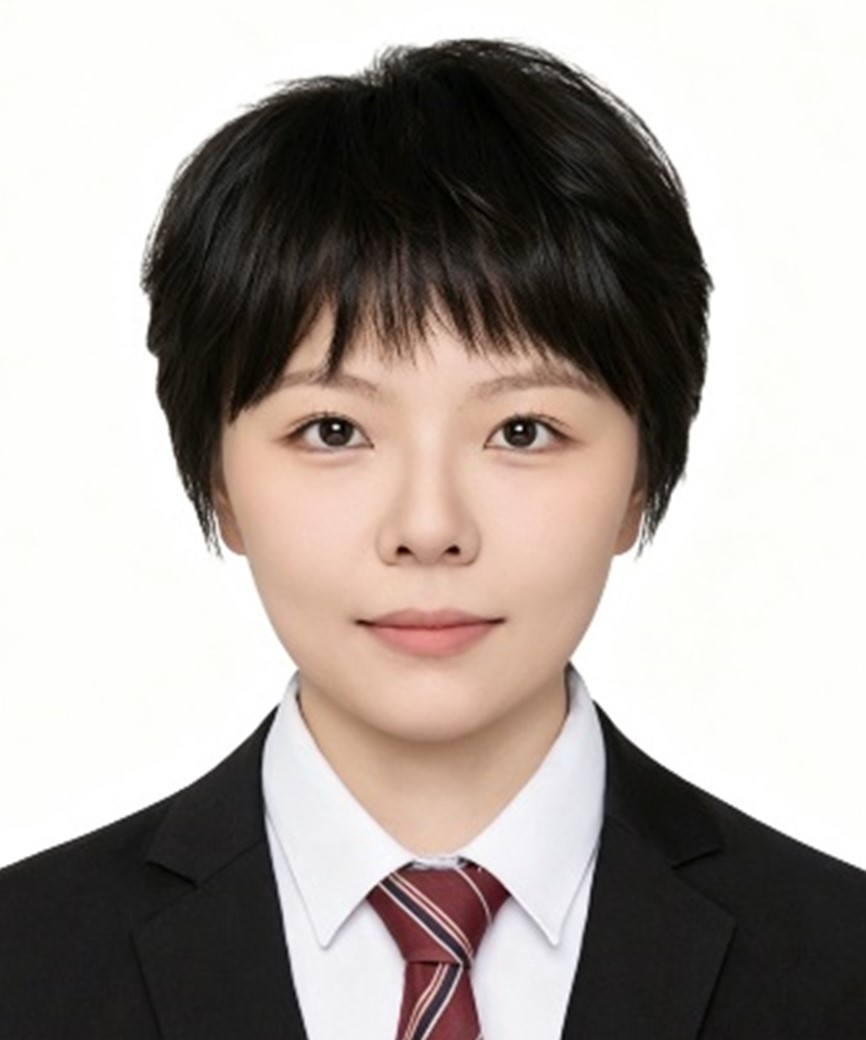}}]{Ruxin Ding} 
received her B.Sc. degree in computer science from the University of Nottingham Ningbo China in 2021. She is currently pursuing a Ph.D. in Computer Science at the University of Nottingham Ningbo China. Her research interests include spatiotemporal feature representation, dynamic texture recognition, and machine learning.
\end{IEEEbiography}

\begin{IEEEbiography}[{\includegraphics[width=1in,height=1.25in,clip]{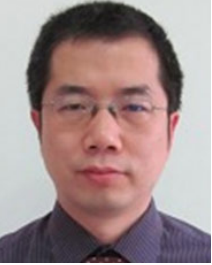}}]{Jianfeng Ren}
(Senior Member, IEEE) received the B.Eng. degree from the National University of Singapore, Singapore, and the M.Sc. and Ph.D. degrees from Nanyang Technological University, Singapore. In 2018, he joined the School of Computer Science, University of Nottingham Ningbo China, where he is now a tenured Associate Professor. He has authored over 110 research papers, including contributions to TIP (3), TIFS (2), TMM (3) and PR (9), and conferences such as CVPR (3), ICCV (1), NeurIPS (1), AAAI (7), ACM MM (3), and IJCAI (1). His research interests include image/video processing, statistical pattern recognition, machine learning, and radar target recognition. 
He serves as an Associate Editor for Visual Computer and IET Biometrics, and as a Guest Editor for Symmetry. He has been a reviewer for more than 20 journals, including TIP and TSMC. He has also served as a TPC Member for CVPR, ICML, NeurIPS, AAAI, IJCAI and ECAI; Area Chair for ACM MM 2024/2025/2026; and Program Chair for ICVRT 2024. 
\end{IEEEbiography}

\begin{IEEEbiography}[{\includegraphics[width=1in,height=1.25in,clip]{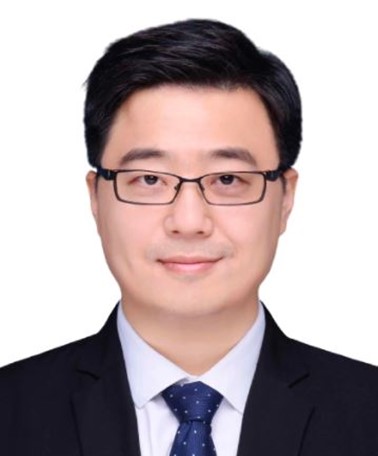}}]{Heng Yu} (M'10–SM'24) 
received his B.Eng. and Ph.D. degrees in Electrical and Computer Engineering from the National University of Singapore (NUS), in 2006 and 2011 respectively. He was a research scientist at the University of Erlangen-Nuremberg, and subsequently a research fellow at NUS. He was an Assistant Professor at the United Arab Emirates University, and briefly a Xinghai Associate Professor at Dalian Maritime University. He is now an Associate Professor in the School of Computer Science, University of Nottingham Ningbo China. His research interests focus on adaptive and reliable embedded systems, as well as AI in embedded computing. His work received the best paper award or nominations at ACM CF'17, FPT'13, and SAFECOMP'12. He served as an Associate Editor for the IEEE Transactions on Circuits and Systems II: Express Briefs. He has been the PI/Co-PI for multiple national/provincial/municipal research grants in China.
\end{IEEEbiography}

\begin{IEEEbiography}[{\includegraphics[width=1in,height=1.25in,clip]{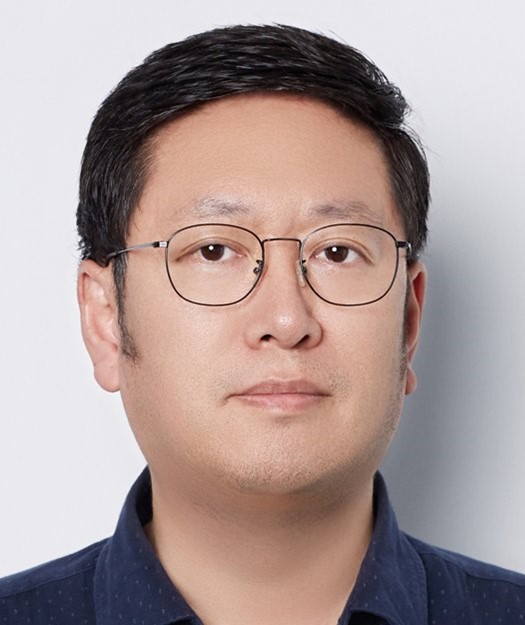}}]{Jiawei Li} 
received his B.Sc. and Ph.D. degrees in engineering from the Harbin Engineering University, Harbin, China, in 1992 and 1998 respectively. He became a lecturer at the Robotics Institute of Harbin Institute of Technology, China, in 1999. Currently, he is an Assistant Professor in the School of Computer Science at University of Nottingham Ningbo China. His research focuses on decision-making methodologies including fuzzy logic, evolutionary game theory, hyper-heuristics, as well as their applications in real-world economic and social systems. He has special interests on the theories of bargaining, game models, and economics.  
\end{IEEEbiography}

\begin{IEEEbiography}[{\includegraphics[width=1in,height=1.25in,clip]{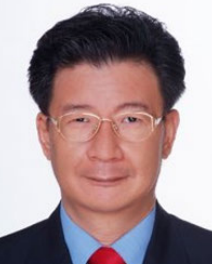}}]{Xudong Jiang}
(Fellow, IEEE) received the B.Eng. and M.Eng. degrees in electrical engineering from the University of Electronic Science and Technology of China (UESTC), Chengdu, China, in 1983 and 1986, respectively, and the Ph.D. degree in electrical engineering from Helmut Schmidt University, Hamburg, Germany, in 1997.,From 1998 to 2004, he was with the Institute for Infocomm Research, A*STAR, Singapore, as a Lead Scientist and the Head of the Biometrics Laboratory. In 2004, he joined Nanyang Technological University (NTU), Singapore, as a Faculty Member, where he served as the Director of the Centre for Information Security from 2005 to 2011. He is currently a Professor with the School of Electrical and Electronic Engineering (EEE), NTU, where he is also the Director of the Centre for Information Sciences and Systems. He holds seven patents and has authored more than 300 articles, where 43 articles were presented in top conferences CVPR/NeurIPS/ICCV/ECCV/AAAI and over 70 articles were published in the IEEE journals, with 14 articles in IEEE Transactions on Pattern Analysis and Machine Intelligence (TPAMI) and 20 articles in IEEE Transactions on Image Processing (TIP). His current research interests include computer vision, machine learning, image processing, pattern recognition, and biometrics.,Dr. Jiang served as an IFS TC Member for the IEEE Signal Processing Society from 2015 to 2017 and an Associate Editor for IEEE Signal Processing Letter from 2014 to 2018 and IEEE Transactions on Image Processing (TIP) from 2016 to 2020. He currently serves as a Senior Area Editor for IEEE TIP and the Editor-in-Chief for IET Biometrics.
\end{IEEEbiography}

\vfill

\end{document}